\documentclass[11pt,a4paper]{article}
\usepackage[utf8]{inputenc}
\usepackage{amsmath}
\usepackage{amsfonts}
\usepackage{amssymb}
\usepackage{amsthm}
\usepackage{makeidx}
\usepackage{amsmath}
\usepackage{graphicx}
\usepackage[outdir=./]{epstopdf}
\usepackage{multirow}
\usepackage[a4paper, total={6.3in, 7.5in}]{geometry}
\usepackage[linesnumbered]{algorithm2e}
\usepackage{matlab-prettifier}
\usepackage{caption}
\usepackage{lscape} 

\usepackage{xcolor}

\theoremstyle{definition}

\theoremstyle{plain}

\theoremstyle{plain}

\theoremstyle{plain}

\theoremstyle{plain}

\theoremstyle{plain}

\providecommand{\keywords}[1]
{
	\small	
	\textbf{\textit{Keywords---}} #1
}

\usepackage[english]{babel}
\usepackage{csquotes}
\usepackage{authblk}
\usepackage{tikz,pgfplots}
\usepackage{lscape} 

\usepackage[backend=bibtex,style=numeric-comp,giveninits=true,natbib]{biblatex}
\bibliography{SVMbib.bib}	
\title{Training very large scale nonlinear SVMs
	using Alternating Direction Method of Multipliers
	coupled with the Hierarchically Semi-Separable kernel approximations}
\author[1]{S. Cipolla \thanks{Email: \texttt{scipolla@ed.ac.uk}}}
\author[1]{J.Gondzio \thanks{Email: \texttt{j.gondzio@ed.ac.uk}} }
\affil[1]{ \footnotesize{The University of Edinburgh, School of Mathematics}}
\date{November 8th, 2021}
\begin{document}
	\maketitle

\begin{abstract}
{Typically, nonlinear Support Vector Machines (SVMs) produce significantly higher classification quality when compared to linear ones but, at the same time, their computational complexity is prohibitive for large-scale datasets: this drawback is essentially related to the necessity to store and manipulate large, dense and unstructured kernel matrices. Despite the fact that at the core of training a SVM there is a \textit{simple} convex optimization problem, the presence of kernel matrices is responsible for dramatic performance reduction, making SVMs unworkably slow for large problems. Aiming to an efficient solution of large-scale nonlinear SVM problems, we propose the use of the \textit{Alternating Direction Method of Multipliers} coupled with \textit{Hierarchically Semi-Separable} (HSS) kernel approximations. As shown in this work, the detailed analysis of the interaction among their algorithmic components unveils  a particularly efficient framework and indeed, the presented experimental results demonstrate a significant speed-up when compared to the \textit{state-of-the-art} nonlinear SVM libraries (without significantly affecting the classification accuracy).}
\end{abstract}

\keywords{Computing science, Support Vector Machines, Hierarchically Semi-Separable kernel approximations, Alternating Direction Method of Multipliers.}

\section{Introduction} \label{sec:Introductio}
{Support vector machine (SVM) is one of the most well-known supervised classification method which has been extensively used in different fields. At its core, training nonlinear SVMs classifier boils down to a solution of a convex Quadratic Programming (QP) problem whose running time heavily depends on the way the  quadratic term  \textit{interacts} with the chosen optimizer. Typically, such interaction, is represented by the solution of a linear system involving the quadratic term (perhaps in some suitably modified version). However, in the nonlinear SVM case, the quadratic term involves a kernel matrix which (except for the linear kernel) is a dense and unstructured matrix.
Solving (or merely storing) a linear system involving such matrices 
may result in unworkably slow algorithms for large scale problems. Although the use of kernel approximations in SVMs classification has been for a long time a relevant research question, see Section \ref{sec:backgorund} for references, the existing structured approximations are not always able to capture the \textit{essential features} of the kernel (see, once more, Section \ref{sec:backgorund} for a detailed explanation of this statement) and, moreover, the selected structure for the kernel approximation may not be  exploitable by the chosen optimizer. Aim of this work is to devise a computational framework based on the use of the \textit{Alternating Direction Method of Multipliers} (ADMM) \cite{boyd2011distributed}  coupled with \textit{Hierarchically Semi-Separable} (HSS)  \cite{MR2191196} kernel approximations. Indeed, the latter choice, if on one hand allows to produce kernel approximations essentially in \textit{a matrix-free} regime  and with guaranteed accuracy \cite{9139856}, on the other, allows the efficient solution of (shifted) linear systems involving it. In turn, when QP problems are solved using ADMM, the solution of shifted kernel linear systems is the main expensive computational task. Such a harmonized interaction between the kernel approximation and the optimizer not only allows a fast training phase but also makes possible a fast grid search for optimal hyperparameters selection through caching the HSS approximation/factorization. }

\subsection{Background} \label{sec:backgorund}

Support vector machines (SVMs) \cite{boser1992training,cortes1995support} are  useful and widely used
classification methods. Training a nonlinear SVM has at its core (in its dual form) the solution of the following convex quadratic optimization problem:
\begin{equation} \label{eq:SVM_problem}
\begin{aligned}
\min_{\mathbf{x} \in \mathbb{R}^d} \;& f(\mathbf{x}):=\frac{1}{2}\mathbf{x}^TY K Y\mathbf{x} -\mathbf{e}^T\mathbf{x} \\ 
\hbox{s.t.} \; & \mathbf{y}^T\mathbf{x} =0, \\
& {x}_i \in [0,C] \hbox{ for all } i=1, \dots,d,
\end{aligned}
\end{equation}	
where $y_{i} \in \{-1,1\}$ are target labels, $Y:= diag(\mathbf{y})$, $K_{ij}:=K(\mathbf{f}_{i},\mathbf{f}_{j})$ is a Positive Definite Kernel \cite[Def. 3]{MR2418654}, $\mathbf{f}_{i} \in \mathbb{R}^r$ are feature vectors  and $\mathbf{e}$ is the vector of all ones. 
 
Once a solution $\bar{\mathbf{x}}$ of problem \eqref{eq:SVM_problem} has been computed, the classification function for an unlabelled data $\mathbf{f}$ can be determined by
\begin{equation*}
	\tilde{y}= sign( \sum_{i=1}^d y_i \bar{x}_i  K(\mathbf{f}_i, \mathbf{f})+b ).
\end{equation*} 
The bias term $b$ is computed using the support vectors that lie on the margins, i.e., considering $j$ s.t. $0 < \bar{x}_j <C$, the following  formula is used:

\begin{equation} \label{eq:bias_formula}
	b = \sum_{i=1}^d y_i \bar{x}_i K(\mathbf{f}_i, \mathbf{f}_j) -y_j.
\end{equation}

Despite their simplicity, when compared with Neural Networks (NNs), 
nonlinear SVMs are still recognised by practitioners of Machine Learning and Data Science as the preferred choice for classification tasks in some situations.  In particular, the community seems to widely agree on the fact that NNs are not efficient on low-dimensional input data because of their huge overparametrisation and, in this case, SVMs may represent the \textit{state of the art} for classification. Indeed, SVMs have only two hyperparameters (say the choice of a kernel-related parameter $h$ and the penalization constant $C$), so they are very easy to tune to specific problems: the parameter tuning is usually  performed by a simple grid-search through the parameter space. 

On the other hand, even if the SVM training is related to  a convex optimization problem for which there exist efficient solution methods, training SVMs for large scale datasets may be a computationally challenging option essentially due to the fact that, in order to be able to use the Kernel Trick, SVMs cache a value for the kernelized ``distance'' between any two pairs of points: for this reason an  $O(d^2)$ storage requirement is to be expected. In general, without any particular \textit{specialization}, training SVMs is unworkably slow for sets beyond, say, $10^4$ datapoints.
 
Without any doubts, the most successful class of methods designed to handle storage difficulties, is represented by decomposition methods \cite{osuna1997training,joachims1998making,platt1998fast,fan2005working}: unlike most optimization methods which update all the variables in each step of an iterative process, decomposition methods modify only a subset of these at every iteration leading, hence, to a small sub-problem to be solved in each iteration. A prominent example in this class is represented by \cite{CC01a} which delivers, somehow, a standard benchmark comparison in the SVMs training panorama. It is important to note at this stage that since only few variables are updated per iteration, for difficult/large-scale problems, decomposition methods may suffer from a slow convergence.

\begin{figure}[ht!]
	\centering
	\includegraphics[trim={0cm 0.15cm 0cm 0cm},width=.3\textwidth]{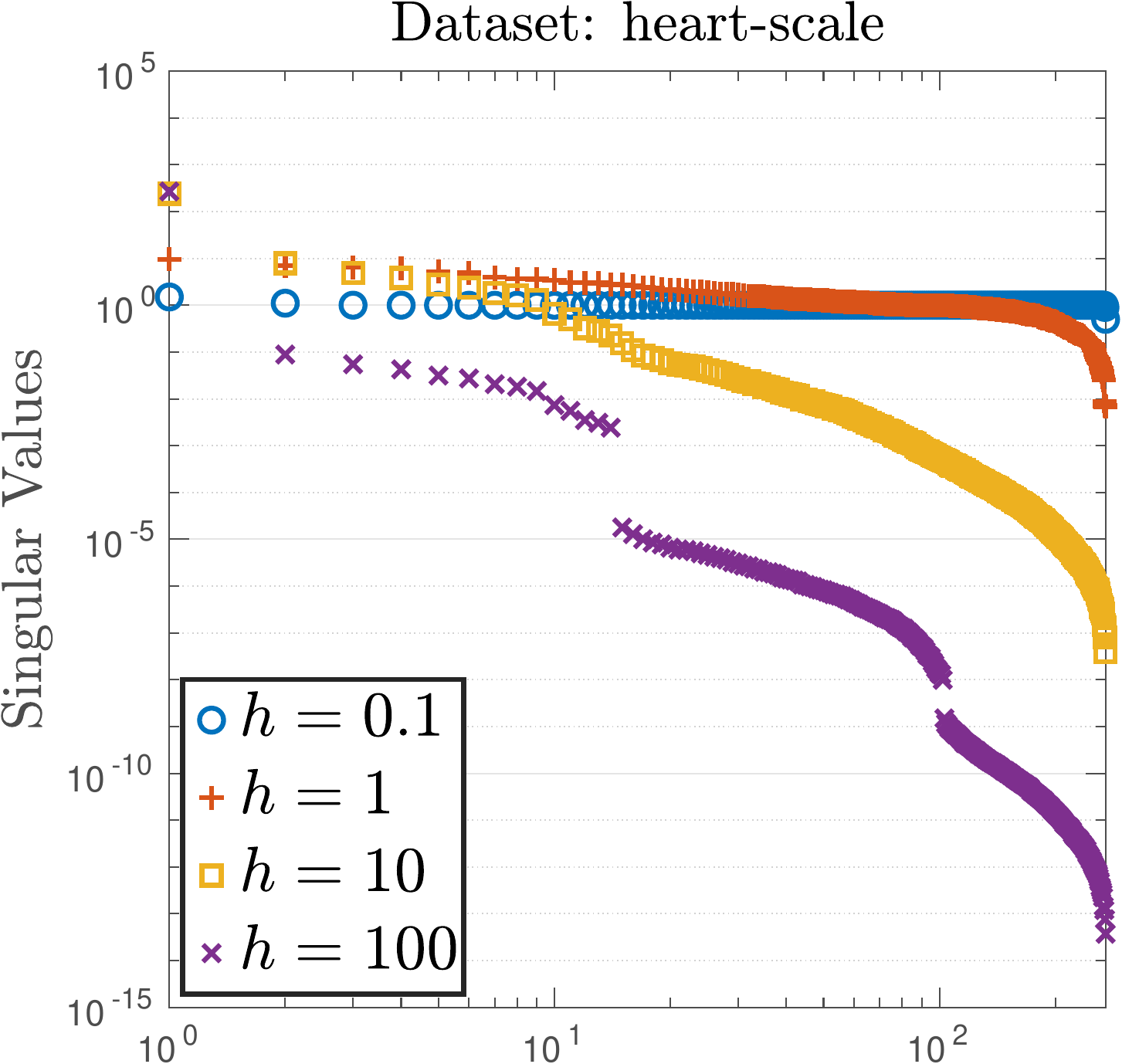}
	\includegraphics[trim={0cm 4cm 0cm 20cm},width=0.69\textwidth]{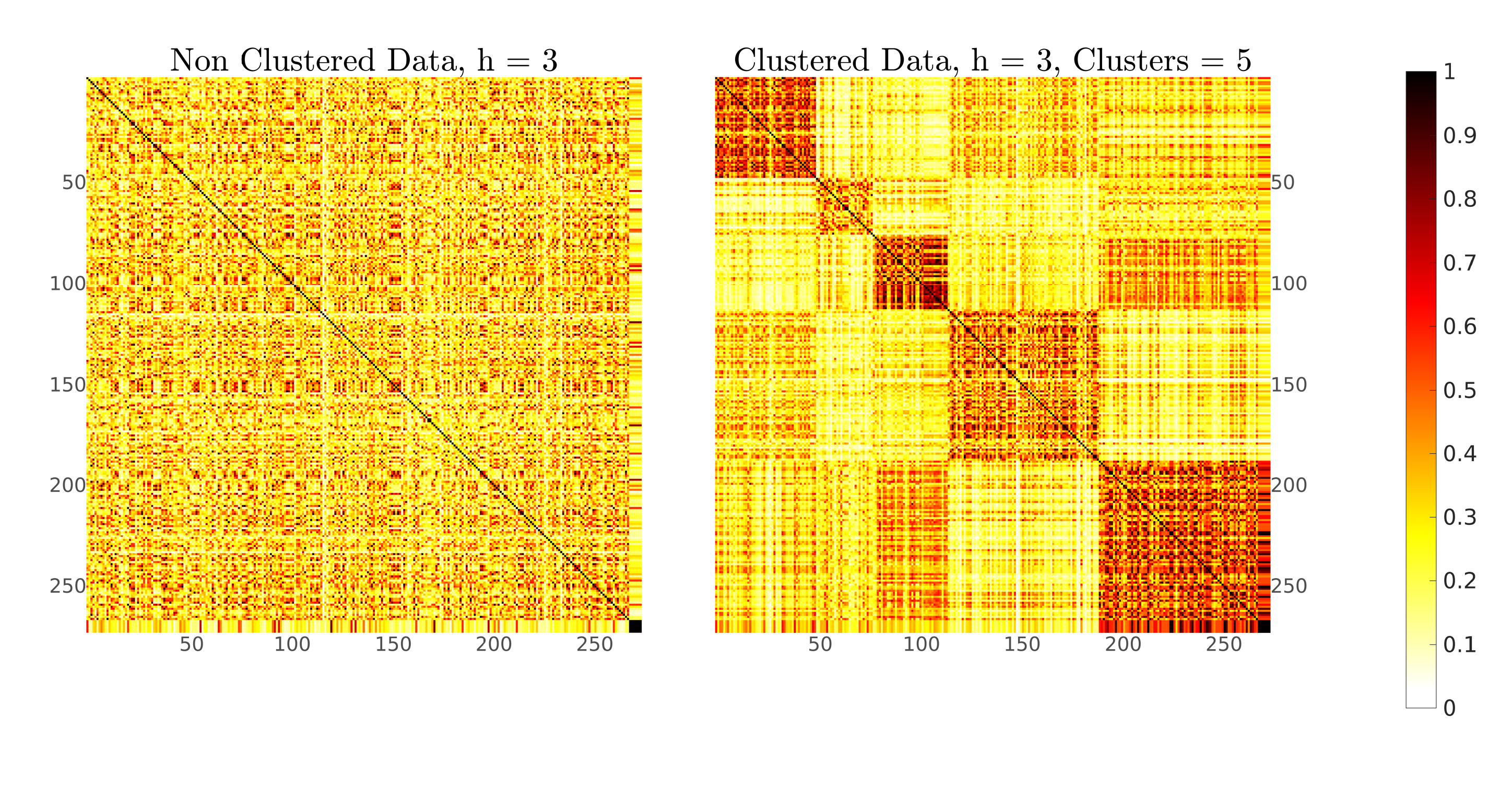}
	\caption{Left Panel: decay of the singular values for Gaussian Kernel matrices. Right Panel: Gaussian Kernel matrices obtained with/without preliminary data clustering. Dataset: \texttt{heart\_scale} \cite{CC01a}. \label{fig:svddecay}}
\end{figure}

On the other hand, an alternative way to overcome storage issues is to approximate the kernel matrix $K$ and, indeed, there is a rich literature concerned with the acceleration of kernel methods which are usually based on the efficient approximation
of the kernel map. The most popular approach is to construct a low-rank matrix approximation of the kernel matrix reducing the arithmetic and storage cost \cite{golub2012matrix,fine2001efficient,mahoney2011randomized,MR2806637,MR2366406,sarlos2006improved,MR2249884,MR3543523,MR3033372,zhang2010clustered}. We mention explicitly Nyström-type methods \cite{williams2001using,MR3543523,MR2930630} and random feature maps to approximate the kernel function directly \cite{rahimi2007random} or as a preconditioner \cite{MR3713904}. However, the numerical rank of the kernel matrix depends
on parameters, which are, in turn, data-dependent: the Eckart–Young–Mirsky theorem, see \cite[Sec. 2.11.1]{MR1442956} justifies low rank approximations only when the kernel matrix is characterized by a sufficiently fast decay of the singular values. For example, the Gaussian kernel matrix, i.e., $K_{ij}=\exp^{-\frac{\|\mathbf{f}_i-\mathbf{f}_i\|^2}{2h^2}}$,  is approximately low-rank only if  $h > 0$  is sufficiently large (see the left panel in Figure \ref{fig:svddecay} for an example) but, for classification purposes, a small value of $h$ may be required.

Several methods were proposed to overcome the fact that $K$ is
not necessarily approximately low-rank. The main idea, in this context, relies on the initial splitting of the data into clusters, so that
between-classes  interactions in the kernel matrix may be represented/well approximated by either sparse or low-rank matrices  \cite{MR3634887,MR4039525, you2018accurate} (see right panel in Figure \ref{fig:svddecay} for a pictorial representation of this idea).

\subsection{Contribution}
{This work represents a methodological contribution for the efficient solution of SVM problems.} {In particular, the aim} of this work is to propose and analyse the use of the \textit{Hierarchically Semi-Separable} (HSS) matrix representation \cite{MR2191196} for the solution of large scale kernel SVMs. Indeed, the use of HSS approximations of kernel matrices has been already investigated in \cite{rebrova2018study,9139856} for the solution of large scale Kernel Regression problems. The main reason for the choice of the HSS structure in this context can be summarised as follows:

\begin{itemize}
	\item using the STRUctured Matrix PACKage (\texttt{STRUMPACK}) \cite{MR3532846} it is possible to obtain HSS approximations of the kernel matrices 
	without the need to store/compute explicitly the whole
	matrix $K$. {Indeed, for kernel matrix approximations, \texttt{STRUMPACK} uses a partially matrix-free strategy (see \cite{9139856}) essentially based on an adaptive randomized clustering and neighboring-based preprocessing of the data: in the preprocessing step employed by \texttt{STRUMPACK},   \textit{clustering} algorithms are employed to find groups of points with large inter-group distances and small intra-group distances. This feature permits to fully exploit  the underlying data geometry to obtain valuable algebraic approximations of the kernel matrix;}
	\item the resulting approximations allow fast approximate
	kernel matrix computations with linear scalability for the computation of matrix-vector products and solution of linear systems.
\end{itemize}

In particular, we trace the main contribution of this work in  unveiling a particularly efficient interaction between the HSS structure and ADMM \cite{boyd2011distributed} in the SVMs case, see Section  \ref{sec:Computational_Framework}. When problem \eqref{eq:SVM_problem} is suitably reformulated in a form exploitable by ADMM, just the solution of one linear system involving the (shifted) kernel matrix is required per ADMM iteration: kernel matrices approximated using the HSS structure allow highly efficient solutions of such linear systems.  Indeed, in this framework, approximating the kernel matrix with an HSS structure ($h$ fixed)  results in a highly efficient optimization phase for a fixed value of $C$ (see Section \ref{sec:Numerical_Results}).  
It is important to note, moreover, that the computational footprint related to the kernel matrix approximation phase is fully justified by the fact that the same approximation can be \textit{reused} for training the model with different values of $C$; this feature makes our proposal particularly attractive when a fine grid is used for the tuning of the penalization parameter $C$. {It is important to note, at this stage, that also the works \cite{ye2011efficient,huang2019hardware} analyse the use of ADMM for SVMs: in \cite{ye2011efficient} ADMM has been used to solve linear SVMs with feature selection 
whereas in \cite{huang2019hardware} a hardware-efficient nonlinear SVM training 
algorithm has been presented in which the Nystr\"{o}m approximation is exploited to reduce the dimension of the kernel matrices. Both works represent and use, somehow, different frameworks and techniques from those presented here.}

%

\section{The computational framework} \label{sec:Computational_Framework}
Problem \eqref{eq:SVM_problem} can be written as follows:
\begin{align} \label{eq:QP_problem_ADMM_IF}
\begin{aligned}
\min_{\mathbf{x},\;  \mathbf{z} \in \mathbb{R}^d} \;& \frac{1}{2}\mathbf{x}^TYKY\mathbf{x} -\mathbf{e}^T\mathbf{x}+I_{\mathbf{y}^T\mathbf{x}=0}(\mathbf{x})+I_{[0,C]}(\mathbf{z})  \\ 
\hbox{s.t.} \; & \mathbf{x}-\mathbf{z}=0 ,\\
\end{aligned}
\end{align}
where, for a given subset $S \subset \mathbb{R}^d$, $I_{S}(\mathbf{x})$ is the indicator function of the set $S$, defined as
\begin{equation*}
I_{S}(\mathbf{x}):= \begin{cases}
0              & \hbox{if } \mathbf{x}\in S\\
+\infty        & \hbox{if } \mathbf{x}\notin S. 
\end{cases}	
\end{equation*}
The Augmented Lagrangian corresponding to \eqref{eq:QP_problem_ADMM_IF} reads as
{
\begin{equation}  \label{eq:Lagrangian}
\mathcal{L}_{\beta}(\mathbf{x}, \mathbf{z}, \boldsymbol{\mu})=\frac{1}{2}\mathbf{x}^TYKY\mathbf{x} -\mathbf{e}^T\mathbf{x}+I_{\mathbf{y}^T\mathbf{x}=0}(\mathbf{x})+I_{[0,C]}(\mathbf{z}) -\boldsymbol{\mu}^T(\mathbf{x} -\mathbf{z}) +\frac{\beta}{2}\|\mathbf{x} -\mathbf{z}\|^2.
\end{equation}
}
Reformulation \eqref{eq:QP_problem_ADMM_IF} with an extra copy of variable $x$ makes it easier to exploit partial separability and facilitates a direct application of ADMM to solve it. Indeed, ADMM \cite{boyd2011distributed} is our choice of an (efficient) solution technique for problem \eqref{eq:QP_problem_ADMM_IF}. 
In Algorithm~\ref{alg:ADMM} we summarize its main steps: 

\begin{algorithm}[hbt!]
	\caption{ADMM}\label{alg:ADMM} 
	\For{$k=0,1, \dots$}{
		$\mathbf{x}^{k+1}=\min_{\mathbf{x} \in  \mathbf{R}^d}  \mathcal{L}_{\beta}(\mathbf{x},\mathbf{z}^{k}, \boldsymbol{\mu}^k)$  \tcc*{$\mathbf{x}$ minimization} \label{algLine:x_min}
		$\mathbf{z}^{k+1}= \min_{\mathbf{z} \in  \mathbf{R}^d}  \mathcal{L}_{\beta}(\mathbf{x}^{k+1}, \mathbf{z}, \boldsymbol{\mu}^{k})$  \tcc*{$\mathbf{z}$ minimization} \label{algLine:z_min}
		$\boldsymbol{\mu}^{k+1}=  \boldsymbol{\mu}^{k}-\beta ( \mathbf{x}^{k+1} -\mathbf{z}^{k+1}) $ \tcc*{Multiplier Update}
	}
\end{algorithm}

\subsection{ADMM details} \label{sec:structure_preserving}
Let us observe that the solution of the problem in Line \ref{algLine:x_min} of Algorithm~\ref{alg:ADMM} is equivalent to the solution of the problem

\begin{equation}\label{eq:constrained_quadratic}
\begin{aligned}
\min_{\mathbf{x} \in \mathbb{R}^d} \;& \frac{1}{2}\mathbf{x}^TY \underbrace{(K+\beta I)}_{=:K_{\beta}} Y\mathbf{x} -\underbrace{(\mathbf{e}+\boldsymbol{\mu}^k+\beta \mathbf{z}^k)^T}_{=:\mathbf{q}^k}\mathbf{x} \\ 
\hbox{s.t.} \; & \mathbf{y}^T\mathbf{x} =0. \\
\end{aligned}
\end{equation}
Writing the KKT conditions of problem \eqref{eq:constrained_quadratic}, i.e.,

\begin{equation*}
	\begin{bmatrix}
	YK_{\beta}Y & - \mathbf{y} \\
	- \mathbf{y}^T & 0
	\end{bmatrix}\begin{bmatrix}
	\mathbf{x}\\
	\lambda
	\end{bmatrix}=\begin{bmatrix}
	\mathbf{e}+\boldsymbol{\mu}^k+\beta \mathbf{z}^k\\
	0
	\end{bmatrix},
\end{equation*}
and eliminating the variable $\lambda$, it is possible to write its solution in a  closed form:
\begin{equation*}
	\mathbf{x}^{k+1}=YK_{\beta}^{-1}Y\mathbf{q}^k-\frac{\mathbf{e}^TK_{\beta}^{-1}Y\mathbf{q}^k}{\mathbf{e}^TK_{\beta}^{-1}\mathbf{e}}YK^{-1}_{\beta}\mathbf{e},
\end{equation*}
where we used the fact that $Y\mathbf{y}=\mathbf{e}$. Moreover, the problem at Line \ref{algLine:z_min} of Algorithm~\ref{alg:ADMM} can be written alternatively as

\begin{align*} 
\arg\min_{\mathbf{z} \in [0,C]} \; g(\mathbf{z}) :=\frac{\beta}{2}\mathbf{z}^T\mathbf{z}  -\beta \mathbf{z}^T  {\mathbf{x}^{k+1}}+\mathbf{z}^T{\boldsymbol{\mu}^{k}},
\end{align*}
which also has a closed-form solution (see \cite[Example 2.2.1]{MR3444832}):

\begin{equation}\label{eq:z_minsolution}
	\mathbf{z}^{k+1}=\Pi_{[0,C]}(\mathbf{x}^{k+1} -\frac{1}{\beta} \boldsymbol{\mu}^{k}),
\end{equation}
where $\Pi_{[0,C]}$ is the component-wise projection onto the interval $[0,C]$. Summarizing the observations carried out in this section, we observe that  Algorithm~\ref{alg:ADMM} can be written  in closed form as in Algorithm~\ref{alg:ADMM_closed form}:

\begin{algorithm}[hbt!]
	\caption{Closed form ADMM for problem \eqref{eq:QP_problem_ADMM_IF}}\label{alg:ADMM_closed form} 
	\For{$k=0,1, \dots$}{
		$\mathbf{x}^{k+1}=YK_{\beta}^{-1}Y\mathbf{q}^k-\frac{\mathbf{e}^TK_{\beta}^{-1}Y\mathbf{q}^k}{\mathbf{e}^TK_{\beta}^{-1}\mathbf{e}}YK^{-1}_{\beta}\mathbf{e}$ \tcc*{$\mathbf{x}$ minimization}     \label{algLine:x_min2}
		$	\mathbf{z}^{k+1}=\Pi_{[0,C]}(\mathbf{x}^{k+1} -\frac{1}{\beta} \boldsymbol{\mu}^{k})$ \tcc*{$\mathbf{z}$ minimization}  \label{algLine:z_min2}
		$\boldsymbol{\mu}^{k+1}=  \boldsymbol{\mu}^{k}-\beta (\mathbf{x}^{k+1}-\mathbf{z}^{k+1}) $ \tcc*{Multiplier Update} \label{algLine:mu_update}
	}
\end{algorithm}

\subsubsection{Computational Cost and Convergence}
Algorithm \ref{alg:ADMM_closed form} requires the solution of a linear system involving the matrix $K_{\beta}$ at every iteration (the vector $YK_{\beta}^{-1}\mathbf{e}$ can be precomputed) plus a series of operations of linear complexity. Moreover, since Algorithm \ref{alg:ADMM_closed form} is a particular instance of ADMM, it is convergent, see \cite{boyd2011distributed}.


%
%

\section{Experiments} \label{sec:experiments}

\subsection{Hierarchically Semi-Separable matrix representation} 

As already pointed out previously, one of the main computational issues associated with problem \eqref{eq:SVM_problem} relates to the fact that the matrix $K$ is usually dense and of large dimension: the cubic computational complexity of application and the quadratic storage requirements for kernel matrices limit the applicability of kernel methods for SVM in large scale applications. To overcome this problem many different approaches have been proposed in literature, see the discussion in Section \ref{sec:Introductio}. The one we decide to take into account here, is the Hierarchically Semi-Separable (HSS) approximation of the kernel matrix in the form proposed in \cite{9139856}. In general, the HSS approximation of a given matrix uses a hierarchical block $2 \times 2$ partitioning of the matrix where all off-diagonal blocks are compressed, or approximated, using a low-rank product \cite{MR2191196}. 
The accurate description of such techniques in the case of kernel matrices is out of the scope of this work and, for this reason, we refer the reader to \cite[Sec. II.B -- II.C]{9139856} for the full details. Instead, for our purposes, we mention explicitly the peculiarities of the approach prosed in \cite{9139856} (HSS-ANN) which have driven our choice:

\begin{itemize}
	\item {instead of using a randomized sampling (see \cite{MR2854612}) to approximate column range of sub-matrices of $K$, this approach uses the kernel function to assess the similarity between data points and hence to identify the dominating entries of the kernel matrix. In particular, the columns corresponding to dominating Approximate Nearest Neighbours (ANN, see \cite{MR3340203,MR3565582}) of the data points  are selected to produce approximations of the column basis of particular sub-matrices of $K$, see \cite[Sec. II.B]{9139856}. As a result, the overall sampling strategy fully exploits the geometry of the underlying data-set.	
}
	\item the overall complexity of the HSS-ANN construction (excluding the preprocessing phase on the data) is $O(r^2d)$ where $r$ is the maximum HSS rank, i.e., the maximum rank over all off-diagonal blocks in the HSS hierarchy, see \cite[Sec. II.C and Alg. 3]{9139856}. The storage complexity of HSS-ANN is $O(dr)$;
	\item after the construction, the (shifted) HSS kernel matrix approximation can be factorized into a $U LV$ form \cite{MR2262971}, where $L$ is lower triangular and $U$ and $V$ are orthogonal. This factored form, computed just once for fixed $h$ in our approach, can be used to solve linear systems involving the (shifted) kernel matrix. 
	
\end{itemize}

\subsection{Implementation details}
In Algorithm \ref{alg:ADMM_STRumpack} we summarise the pseudo-code of our implementation. It is based on \texttt{STRUMPACK} library (\texttt{Version 5.1.0})  \cite{STRWS, ghysels2017robust}, which provides efficient routines for the approximation $\tilde{K}$  of  a kernel matrix $K$ (see Line \ref{algLine:compression} of Algorithm \ref{alg:ADMM_STRumpack}). Moreover, once $\tilde{K}$ is obtained, it provides efficient routines for the solution of linear systems of the form $\tilde{K}_{\beta}\mathbf{x}=\mathbf{b}$ through the exploitation of a $ULV$ factorization (see Line \ref{algLine:factorization} of Algorithm \ref{alg:ADMM_STRumpack}). It is worth noting that for a fixed kernel value $h$ the approximation $\tilde{K}$ and the factorization $ULV$ of $\tilde{K}_{\beta}$ are computed just once and then \textit{reused} for all the values $C$ in the grid search.

It is also important to note that in practice the bias $b$ is  obtained averaging over  all the support vectors that lie on the margin {instead of using equation \eqref{eq:bias_formula}.} {Indeed, defining $M:=\{j \; | \;  0 < \bar{x}_j <C\}$ and $ \bar{{e}}_j=1$ if $j \in M$ or  $ \bar{{e}}_j=0$} otherwise, the bias $b$ is often computed using 
\begin{equation} \label{eq:bias_formula_Average}
{b = \frac{1}{|M|}\sum_{{j\,\in M}}(\sum_{i=1}^d y_i \bar{x}_i K(\mathbf{f}_i, \mathbf{f}_j) -y_j) = \frac{1}{|M|} (\bar{\mathbf{x}}_{\mathbf{y}}^TK \bar{\mathbf{e}} -  \sum_{j\,\in M} y_j)},
\end{equation}
where $(\bar{\mathbf{x}}_{\mathbf{y}})_j:= y_j \bar{{x}}_j $. If the full kernel matrix $K$ is not available, computing \eqref{eq:bias_formula_Average} may be time consuming for large datasets since it requires a series of kernel evaluations. On the other hand, the right-hand side of equation \eqref{eq:bias_formula_Average} suggests that if an approximation $\tilde{K}$ of $K$ is available for which matrix vector products can be \textit{inexpensively} evaluated, the bias computation requires exactly just one matrix vector product and  one scalar product. This is indeed the case when an HSS approximation of the kernel matrix is available and we exploit this property in our implementation, see Line \ref{algLine:bias_computation} in Algorithm \ref{alg:ADMM_STRumpack}.

\begin{algorithm}[hbt!]
	\caption{SVM training/testing using Strumpack and ADMM}\label{alg:ADMM_STRumpack} 
	\KwIn{$K$ kernel function,  $h$ kernel parameter, $\beta$ ADMM parameter, $ F_{train} \in \mathbb{R}^{r \times d}$, $\mathbf{y}_{train} \in \mathbb{R}^{d}$,  ${F_{test}} \in \mathbb{R}^{r \times m}$ , $\mathbf{y}_{test} \in \mathbb{R}^{m}$ -- training and testing data. }
	$\tilde{K}=\mathtt{HSScompression}(K(F_{train}, F_{train}), h)$ \label{algLine:compression}  \;
	$\tilde{K}_\beta = \tilde{K}+\beta I  $\;
	$[U, L, V ] = \mathtt{ULVfactorization}(\tilde{K}_\beta)$ \label{algLine:factorization} \;
	$\mathbf{w} = (ULV)^{-1}\mathbf{e}$\;
	$w_1 = \mathbf{e}^T \mathbf{w}  $\;
	$\mathbf{w} = Y_{train}\mathbf{w}$ \;
	\For{$C \in \{ C_1, \dots, C_{max}\}$}{
		Initialize $\mathbf{x}_0, \mathbf{z}_0, \boldsymbol{\mu}_0$ \;
	\For{$k=0,1, \dots, MaxIt$}{
		$w_2 = \mathbf{w}^T\mathbf{x}^k$\;
		$\mathbf{x}^{k+1}=Y(ULV)^{-1}Y \mathbf{x}^k- \frac{w2}{w1}\mathbf{w} $ \tcc*{$\mathbf{x}$ minimization}     \label{algLine:x_min2_str}
		$	\mathbf{z}^{k+1}=\Pi_{[0,C]}(\mathbf{x}^{k+1} -\frac{1}{\beta} \boldsymbol{\mu}^{k})$ \tcc*{$\mathbf{z}$ minimization}  \label{algLine:z_min2_str}
		$\boldsymbol{\mu}^{k+1}=  \boldsymbol{\mu}^{k}-\beta (\mathbf{x}^{k+1}-\mathbf{z}^{k+1}) $ \tcc*{Multiplier Update} \label{algLine:mu_update_str}
	}
 
 Define $\mathbf{z}_{\mathbf{y}}= Y_{train}\mathbf{z}^{MaxIt} $   \tcc*{Computing Bias} 
Define $\bar{\mathbf{e}}_j = 1$ if $0<(\mathbf{z}_{MaxIt})_j<C$ or $\bar{\mathbf{e}}_j = 0$ otherwise \;  

{$b = \frac{1}{\|\bar{\mathbf{e}}\|_1} ( {\mathbf{z}_{\mathbf{y} } }^T \tilde{K} \bar{\mathbf{e}} - \sum_{j\,:\bar{\mathbf{e}}_j \neq 0} ({\mathbf{y}_{train}})_j) $} \label{algLine:bias_computation} \;

\For{$j=1, \dots, m$}{$
	({\tilde{\mathbf{y}}}_{test})_j= sign( \sum_{i=1}^d (\mathbf{z}_{\mathbf{y}})_i  K((\mathbf{f}_{train})_i, ({\mathbf{f}_{test}})_j)+b )$ \tcc*{Label Assignement} 
	}

}
\end{algorithm}

Finally, to conclude this section, we address briefly the problem of  relating the solution $\tilde{\mathbf{x}}$ of the approximated SVM problem

\begin{equation} \label{eq:SVM_problem_approx}
\begin{aligned}
\min_{\mathbf{x} \in \mathbb{R}^d} \;& \tilde{f}(\mathbf{x}):=\frac{1}{2}\mathbf{x}^TY \tilde{K} Y\mathbf{x} -\mathbf{e}^T\mathbf{x} \\ 
\hbox{s.t.} \; & \mathbf{y}^T\mathbf{x} =0, \\
& {x}_i \in [0,C] \hbox{ for all } i=1, \dots,d,
\end{aligned}
\end{equation}
to the solution $\bar{\mathbf{x}}$  of the original problem \eqref{eq:SVM_problem}.  Indeed, using a similar technique to the one presented in \cite[Sec. 4.1.]{fine2001efficient}, for any unitary invariant form we obtain

\begin{equation}\label{eq:theoretical_bound}
\begin{split}
 |f(\bar{\mathbf{x}})-\tilde{f}(\tilde{\mathbf{x}})| & \leq \max \{ \frac{1}{2}|\tilde{\mathbf{x}}^TY(\tilde{K}-K)Y\tilde{\mathbf{x}}|, \frac{1}{2}|{\bar{\mathbf{x}}}^TY(K-\tilde{K})Y{\bar{\mathbf{x}}}|  \}  \\
& \leq \frac{1}{2}\max\{ \|\tilde{\mathbf{x}}\|^2, \|\bar{\mathbf{x}}\|^2 \}\| \tilde{K}-K\|.
\end{split} 
\end{equation}
 Using the boundedness of $\frac{1}{2}\max\{ \|\tilde{\mathbf{x}}\|^2, \|\bar{\mathbf{x}}\|^2 \}$, we obtain that for $\tilde{K} \to K$ it holds $ \tilde{f}(\tilde{\mathbf{x}}) \to f(\bar{\mathbf{x}})$. Equation \eqref{eq:theoretical_bound} suggests that for increasingly accurate approximations $\tilde{K}$ of $K$, the accuracy classification performance of the \textit{approximate} SVM classifier  \eqref{eq:SVM_problem_approx}  matches increasingly closely the accuracy classification performance of the \textit{exact} SVM classifier  \eqref{eq:SVM_problem}. Nonetheless, we will show experimentally, that this may be also true when quite poor approximations are used, see Table \ref{table:A+S1} in the following section. {Indeed, surprisingly enough, it has been observed multiple times that for kernel methods even  poor approximations of the kernel can suffice to achieve near-optimal performance \cite{rudi2015less,bach2013sharp}.} {On the other hand, it is also important to note that if the matrix $K$ has the \textit{HSS property} (see \cite[Sec. 3]{MR2854612}), and assuming that $\tilde{K}$ is computed by truncating every HSS block with a truncation tolerance $O(\varepsilon)$,  then also the global error $\|K-\tilde{K}\|$  stays of order $O(\varepsilon)$: specific results are available for the Frobenius and spectral norms, see \cite[Corollary 4.3]{MR3152737} and \cite[Theorem 4.7]{MR3928351}. In particular, supposing that every HSS block of $K$ can be approximated with an error $O(\varepsilon)$ by a matrix of rank $r$, if a randomized sampling procedure is used to produce the low rank approximations with oversampling parameter $p$, then $\tilde{K}$ is a global $O(\varepsilon)$ approximation with probability at least $1-6p^{-p}$ (see \cite[Sec. 2.3]{MR2854612} and discussion in \cite[Sec. 3.2]{MR4082285}).}

\subsection{Numerical results} \label{sec:Numerical_Results}
Our code is written in \texttt{C++} and the numerical experiments are 
performed on a Dell PowerEdge R920 machine running Scientific Linux 7 and equipped with Four Intel Xeon E7-4830 v2 2.2GHz, 20M Cache, 7.2 GT/s QPI, Turbo (4x10Cores) 256 GB RAM. 

\begin{table}[h!]
	\begin{tabular}{llllll}
		\hline
		Dataset            & Features & Training Set Dim. & $|{Train}_{+}|$ & Test Set Dim. &  $|{Test}_{+}|$ \\ \hline
		a8a                &    122    & 22696 &      5506       & 9865          &   2335    \\ 
		w7a                &    300   & 24692   &    740      & 25057        &     739   \\ 
		rcv1.binary                &    47236   & 20242  &    10491        & 135480       &     71326    \\ 
		a9a                &   122    & 32561   &   7841       & 16281        &     3846    \\ 
		w8a                &   300    & 49749     &     1479   & 14951            &   454 \\ 
		ijcnn1             &    22    & 49990    &     4853    & 91701        &     8712   \\ 
		cod.rna           &    8     & 59535       &    19845  & 271617        &   90539  \\ 
		skin.nonskin*     &    3     & 171540     &  135986     & 73517          &   58212     \\
		webspam.uni*     &    254     & 245000    &  148717       & 105000           &   63472    \\ 
		susy*            &  18 & 3500000 &  1601659 & 1500000 & 686168
	\end{tabular}
	\caption{Problem Set Details.  * = Test Set obtained using  Random 30\%  of the original Training~Set.}
	\label{tab:problem_set}
\end{table}

Table \ref{tab:problem_set} summarizes the details for the chosen dataset. In Tables \ref{table:A+S1} and \ref{table:A+S3} we report the results obtained using our proposal for different parameters related to the accuracy of the HSS-ANN approximation (increasing accuracy) where all the other non specified  HSS-ANN parameters have to be considered the default ones. 
In our experiments we choose, in Algorithm \ref{alg:ADMM_STRumpack}, $MaxIt=10$ and the Gaussian Kernel function $K(\mathbf{f}_i,\mathbf{f}_j)=\exp^{-\frac{\|\mathbf{f}_i-\mathbf{f}_i\|^2}{2h^2}}$. Indeed, it is important to observe that the choice of making a prescribed number of ADMM iterations instead of using a standard stopping criterion, is motivated by the fact that for machine learning applications going for accurate optimal solution does not necessarily have to deliver the best classification accuracy. On the other hand, the fact that one choice of the ADMM parameter $MaxIt$ permits to obtain satisfactory classification accuracy for all the problems in our dataset confirms the robustness of our approach (also if we should mention the experimental observation concerning the fact that for particular problems, a different choice of this parameter may led to better classification performance). Finally, concerning the choice of the ADMM parameter $\beta$, we observed that for larger problems an increasing value of $\beta$ is recommended: we chose $\beta =10^2$ if the training size $d \in [10^4, 10^5]$, $\beta =10^3$ if $d \in [10^5, 10^6]$ and $\beta =10^4$ if $d \geq 10^6$.   

In Table \ref{table:LIBSVM}  we report the results obtained using \texttt{LIBSVM Version 3.25} \cite{CC01a}, which implements specialized  algorithms to address the SVM problem (LIBSVM uses a Sequential Minimal Optimization type decomposition method \cite{bottou2007support,MR2249875,platt1998fast}). In Table \ref{table:rqcqp} we report the results obtained using RACQP \cite{mihic2019managing} (where a multi-block generalization of ADMM is employed, see also \cite{MR4066996,cipolla2020admm} for related theoretical analysis).

In particular, the kernel  parameter $h$  and the penalization term $C$ were estimated by running a grid-check when instances are solved using our proposal (the HSS-ANN accuracy parameters used are those specified in Table \ref{table:A+S3} since our proposal achieves (generally) the best classification accuracy in this case).  Those pairs were then used to solve the instances with LIBSVM and RACQP. The pairs were chosen from a relatively coarse grid, $h, \; C \in \{0.1, 1, 10\}$ because the goal of this experiment is to demonstrate that although our approach uses kernel \textit{approximations}, it can still achieve comparable classification accuracy but with a reduced runtime when compared with other algorithms for the solution of SVM problems which use the \textit{true} kernel matrices.  

\begin{figure}[ht!]
	\centering
	\includegraphics[width=0.30\textwidth]{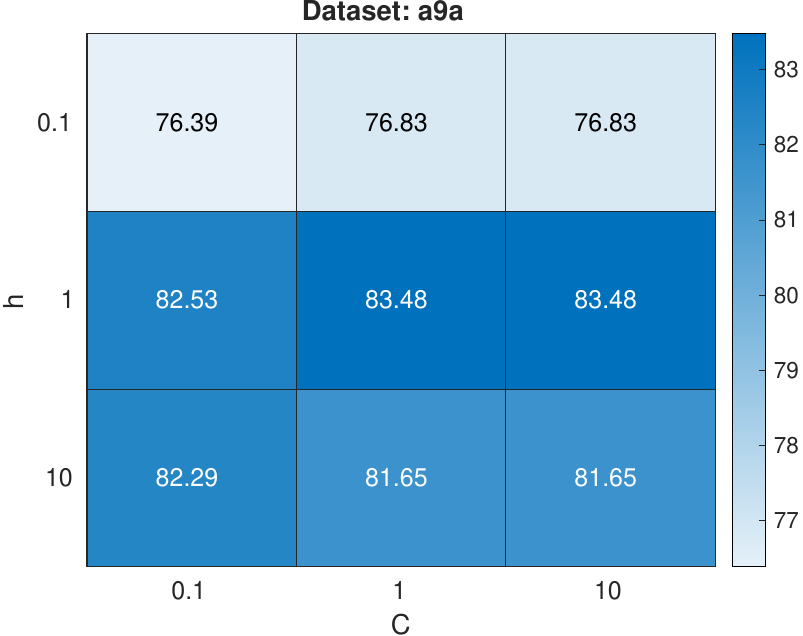}
	\includegraphics[width=0.30\textwidth]{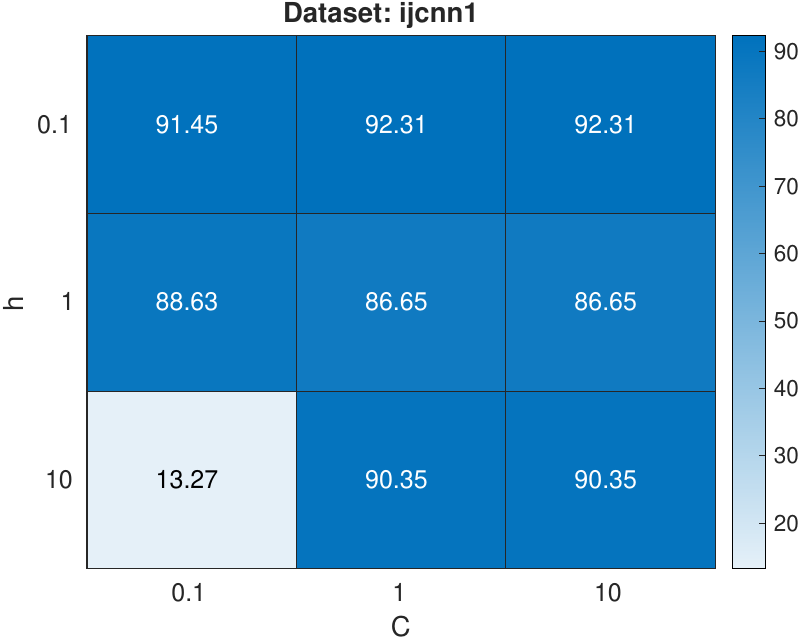}
	\caption{Heatmap of the classification accuracy for the datasets {\texttt{a9a} and \texttt{ijcnn1}}. \label{fig:heatmap}}
\end{figure}

The first important observation concerning  Tables \ref{table:A+S1} and \ref{table:A+S3} is that, unexpectedly (see equation \eqref{eq:theoretical_bound}), increasing the HSS accuracy parameters
(generally) does not lead to a significant increase of classification accuracy: we obtain quite good classification accuracy despite using very rough approximations (see Table \ref{table:A+S1}). 
The problem which benefited most an improved kernel approximation is \texttt{webspam.uni}. 
Indeed, the classification accuracy has increased by nearly $1\%$ in this case. 
At the same time, increasing the HSS accuracy parameters adversely affects the \texttt{Compression} and \texttt{Factorization} time. It is important to note also that the \texttt{ADMM Time} needed to train the model is completely negligible when compared to the time needed to produce the HSS-ANN approximations. As was already pointed out, this feature allows for a very fast grid-search on the parameter $C$ (for the largest considered problem it takes roughly $10s$ to train the model for a fixed $C$).  Indeed, the choice of the parameter $C$ may greatly affect the performance of the classification accuracy (see Figure \ref{fig:heatmap} for some examples).

Concerning the comparison of our approach with LIBSVM and RACQP (compare Tables \ref{table:A+S1} and \ref{table:A+S3} with Tables \ref{table:LIBSVM} and \ref{table:rqcqp}, respectively) several remarks are in order. The first one concerns the coherence of the HSS-ANN approximations with the classification accuracy: the accuracy results obtained for the grid-selected $h$ and $C$ are always comparable to those obtained using LIBSVM and generally better than those obtained using RACQP (both approaches use, in different ways, the true kernel matrices). The second one concerns the computational time: for smallest problems or problems with high dimensional features, our proposal may not be the best performer (see, e.g., the problems \texttt{w7a}, \texttt{rcv1.binary} and \texttt{w8a}) but, on the contrary, when the dimension of the training set increases and the number of features is \textit{small}, the approach proposed in this paper becomes a clear winner (see problems \texttt{ijcnn}, \texttt{cod.rna},  \texttt{webspam.uni} and \texttt{susy}): the goodness and advantages of our approach are further underpinned observing that  the \textit{total training time} needed for the grid search on the parameter $C$ ($h$ fixed) can be roughly obtained multiplying the values in the column \texttt{ADMM Time} by the number of grid values selected for $C$ (in our case $3$). This is not true for LIBSVM and RACQP where the training phase is \textit{restarted from scratch} for all the values $C$ (considering also in this case $h$ fixed).  

Finally, for the sake of fairness, concerning the comparison of running times of our proposal with those from RACQP, we should mention the fact that RACQP is implemented in Matlab, which is presumably slower than a compiled language such as \texttt{C++}. 

\begin{table}[]
	\footnotesize
	\begin{minipage}{.45\linewidth}
		\begin{tabular}{lll}
			\hline
			\multicolumn{1}{|l|}{Dataset} & \multicolumn{1}{l|}{Runtime [s]} & \multicolumn{1}{l|}{Accuracy [\%]} \\ \hline
			a8a                           & 123.308                       & 83.953                        \\
			w7a                           & 148.110                       & 97.904                       \\
			rcv1.binary                  & 261.399                       & 93.247                        \\
			a9a                           & 305.913                       & 82.697                        \\
			w8a                           & 508.232                       & 99.444                        \\
			ijcnn1                        & 345.805                       & 96.007                        \\
			cod.rna                       & 110.997                       & 90.374                        \\
			skin.nonskin                  & 344.938                       & 99.960                         \\
			webspam.uni                   & 13354.384                        & 99.081     \\
			susy                   & $\dagger \dagger$                        &                  
		\end{tabular}
		\caption{LIBSVM. $\dagger \dagger = $ stopped after $10$h.}\label{table:LIBSVM}
	\end{minipage} \hspace{0.5cm} \begin{minipage}{.45\linewidth}
		\begin{tabular}{lll}
			\hline
			\multicolumn{1}{|l|}{Dataset} & \multicolumn{1}{l|}{Runtime [s]} & \multicolumn{1}{l|}{Accuracy [\%]} \\ \hline
			a8a                           &  98.269                       & 79.757                        \\
			w7a                           & 82.838                       & 97.050                        \\
			rcv1.binary                  & 67.830                       & 71.987                        \\
			a9a                           & 206.527                       & 82.237                        \\
			w8a                           & 348.122                      & 97.806                        \\
			ijcnn1                        & 427.551                       & 91.460                        \\
			cod.rna                       & 531.787                       & 33.333                        \\
			skin.nonskin                  & 4689.815                       &  97.649                         \\
			webspam.uni                   & 21669.329                        &  92.830  \\
			susy                   & $\dagger \dagger$                        &                          
		\end{tabular}
		\caption{RACQP. $\dagger \dagger = $ stopped after $10$h.}\label{table:rqcqp}
	\end{minipage}
\end{table} 

\section{Conclusions}

In this work we proposed an ADMM-based scheme (see Algorithm \ref{alg:ADMM_STRumpack}) which employs HSS-ANN approximations (see \cite{9139856} and Section \ref{sec:experiments}) to train SVMs. Numerical experiments obtained using \texttt{STRUMPACK} \cite{STRWS} in a sequential architecture, show that our proposal compares favourably with \texttt{LIBSVM} \cite{CC01a} and \texttt{RACQP} \cite{mihic2019managing} in terms of computational time and classification accuracy when the dimension of the training set increases. Indeed, both \texttt{LIBSVM} and \texttt{RACQP} use different decomposition methods for the \textit{exact} kernel matrix, which may be slow for large scale problems.  Our proposal, instead, resorting on an \textit{all-at-once optimal} exploitation of structured approximations of the kernel matrices, is less prone to the \textit{curse of dimensionality} allowing us to train datasets of larger dimensions. 

\vspace{0.5cm}

{\small
\noindent \textbf{Acknowledgements}: This work was supported by Oracle Labs.
}

\begin{landscape}
{
\begin{table}[]
	\footnotesize
	\begin{tabular}{llllllll}
		\hline
		\multicolumn{1}{|l|}{\multirow{2}{*}{Dataset}} & \multicolumn{3}{l|}{HSS Construction}                                                                                        & \multicolumn{1}{l|}{\multirow{2}{*}{ADMM Time {[}s{]}}} & \multicolumn{2}{l|}{Best Parameters}                & \multicolumn{1}{l|}{\multirow{2}{*}{Accuracy {[}\%{]}}} \\ \cline{2-4} \cline{6-7}
		\multicolumn{1}{|l|}{}                         & \multicolumn{1}{l|}{Compression {[}s{]}} & \multicolumn{1}{l|}{Factorization {[}s{]}} & \multicolumn{1}{l|}{Memory {[}MB{]}} & \multicolumn{1}{l|}{}                                       & \multicolumn{1}{l|}{$h$} & \multicolumn{1}{l|}{$C$} & \multicolumn{1}{l|}{}                                   \\ \hline
		a8a           & 135.923   & 6.181   & 112.968  & 0.300  & 1   & 1,10  & 83.314  \\
		w7a           & 2161.920  & 14.442 & 99.345   & 0.486  & 1   & 1,10  & 97.465  \\
		rcv1.binary   & 6319.780   & 1.665   & 58.839 & 0.173  & 10  & 1,10  & 89.940   \\
		a9a           & 256.032  &  8.162  & 179.192  & 0.471   & 1   & 1,10  & 83.477  \\
		w8a           & 10476.200  & 107.71 & 273.1    & 1.498  & 1   & 1,10  & 97.679  \\
		ijcnn         & 9.772   & 1.980  & 153.586  & 0.470  & 0.1 & 1,10  & 92.403  \\
		cod.rna       & 2.900   & 2.863  & 181.47   & 0.444  & 10  & 0.1     & 89.305  \\
		skin.nonskin & 1127.79  & 11.078 & 538.349  & 1.219 & 10  & 0.1,1,10  & 99.846  \\
		webspam.uni  & 5809.6   & 3.228 & 757.969  & 0.909 &  0.1    & 0.1,1,10  & 95.551 \\
		susy  & 3938.68   & 25.614 & 13599.4  & 9.471 &  1    & 0.1,1,10  & 72.338                                                 
	\end{tabular}
	\caption{Strumpack\&ADMM. Strumpack parameters: \texttt{hss\_rel\_tol}= 1, \texttt{hss\_abs\_tol}= 0.1,  \texttt{hss\_max\_rank}= 200, \texttt{hss\_approximate\_neighbors}= 64. } \label{table:A+S1}

\vspace{0.2cm}

	\begin{tabular}{llllllll}
		\hline
		\multicolumn{1}{|l|}{\multirow{2}{*}{Dataset}} & \multicolumn{3}{l|}{HSS Construction}                                                                                        & \multicolumn{1}{l|}{\multirow{2}{*}{ADMM Time {[}s{]}}} & \multicolumn{2}{l|}{Best Parameters}                & \multicolumn{1}{l|}{\multirow{2}{*}{Accuracy {[}\%{]}}} \\ \cline{2-4} \cline{6-7}
		\multicolumn{1}{|l|}{}                         & \multicolumn{1}{l|}{Compression {[}s{]}} & \multicolumn{1}{l|}{Factorization {[}s{]}} & \multicolumn{1}{l|}{Memory {[}MB{]}} & \multicolumn{1}{l|}{}                                       & \multicolumn{1}{l|}{$h$} & \multicolumn{1}{l|}{$C$} & \multicolumn{1}{l|}{}                                   \\ \hline
		a8a           & 795.597 & 16.276  & 218.673 & 0.588 & 1   & 1,10  & 83.476 \\
		w7a           & 2311.330 & 15.229   & 107.393 & 0.621 & 1   & 1,10     & 97.465 \\
		rcv1.binary   & 14211.0 & 1.425    & 58.84   & 0.210 & 10  & 1,10  & 87.921 \\
		a9a           & 1176.99 & 21.3909  & 379.852 & 0.986  & 1   & 1,10  & 83.643 \\
		w8a           & 10774.900   & 124.076  & 296.472 & 1.738  & 1   & 1,10  & 97.672 \\
		ijcnn         & 21.393 & 2.041  & 168.007 & 0.298 & 0.1 & 1,10  & 92.314 \\
		cod.rna       & 23.242 & 2.377  & 182.424 & 0.280 & 10  & 1,10  & 89.308 \\
		skin.nonskin &  1232.730 & 7.560  & 544.544  & 0.972  & 10  & 0.1,1,10     & 99.855  \\
		webspam.uni  & 7003.52 & 5.640 & 861.542  & 1.297  & 0.1 & 0.1,1,10  & 96.123  \\
		susy    &   14495.9  & 159.972
		& 18264.2  &  15.889 & 1   & 0.1,1,10   &  72.047                                               
	\end{tabular}
	\caption{Strumpack\&ADMM. Strumpack parameters: \texttt{hss\_rel\_tol}= 0.5, \texttt{hss\_abs\_tol}= 0.05,  \texttt{hss\_max\_rank}=~2000, \texttt{hss\_approximate\_neighbors}= 512. } \label{table:A+S3}
\end{table} 
}
\end{landscape}

\printbibliography	
\end{document}